\documentclass[runningheads]{llncs}

\usepackage{graphicx}
\usepackage{amsmath,amssymb} 
\usepackage{tikz}
\usepackage{comment}
\usepackage{color}

\usepackage{pifont}
\newcommand{\crx}[0]{\textcolor{red}{\ding{55}}}

\newcommand{\cbt}[0]{\textcolor{blue}{\ding{51}}}

\usepackage{xspace}
\newcommand{\etal}{\textit{et al}.\xspace}
\newcommand{\ie}{\textit{i}.\textit{e}.\xspace}

\usepackage[nolist]{acronym}
\begin{acronym}[PHOENIX14\textbf{T} ]

\acrodefplural{rnn}[RNNs]{Recurrent Neural Networks}
\acrodefplural{cnn}[CNNs]{Convolutional Neural Networks}
\acrodefplural{hmm}[HMMs]{Hidden Markov Models}
\acrodefplural{gru}[GRUs]{Gated Recurrent Units}
\acrodefplural{crf}[CRFs]{Conditional Random Fields}
\acrodefplural{gan}[GANs]{Generative Adversarial Networks}
\acrodefplural{lcs}[LCSes]{Longest Common Subsequences}
\acrodefplural{gpu}[GPUs]{Graphic Processing Units}
\acrodefplural{slrt}[SLRTs]{Sign Language Recognition Transformers}
\acrodefplural{sltt}[SLTTs]{Sign Language Translation Transformers}

\acro{asl}[ASL]{American Sign Language}
\acro{bsl}[BSL]{British Sign Language}
\acro{bleu}[BLEU]{Bilingual Evaluation Understudy}
\acro{blstm}[BLSTM]{Bidirectional Long Short-Term Memory}
\acro{cnn}[CNN]{Convolutional Neural Network}
\acro{crf}[CRF]{Conditional Random Field}
\acro{cslr}[CSLR]{Continuous Sign Language Recognition}
\acro{ctc}[CTC]{Connectionist Temporal Classification}
\acro{dl}[DL]{Deep Learning}
\acro{dgs}[DGS]{German Sign Language - Deutsche Gebärdensprache}
\acro{dsgs}[DSGS]{Swiss German Sign Language - Deutschschweizer Geb\"ardensprache}
\acro{fc}[FC]{Fully Connected}
\acro{gan}[GAN]{Generative Adversarial Network}
\acro{gpu}[GPU]{Graphics Processing Unit}
\acro{gru}[GRU]{Gated Recurrent Unit}
\acro{hmm}[HMM]{Hidden Markov Model}
\acro{hog}[HOG]{Histograms of Oriented Gradients}
\acro{ip}[IP]{Inner Product}
\acro{isl}[ISL]{Irish Sign Language}
\acro{lcs}[LCS]{Longest Common Subsequence}
\acro{lstm}[LSTM]{Long Short-Term Memory}
\acro{nmt}[NMT]{Neural Machine Translation}
\acro{ocr}[OCR]{Optical Character Recognition}
\acro{ph12}[PHOENIX12]{RWTH-PHOENIX-Weather-2012}
\acro{ph14}[PHOENIX14]{RWTH-PHOENIX-Weather-2014}
\acro{ph14t}[PHOENIX14\textbf{T}]{RWTH-PHOENIX-Weather-2014\textbf{T}}
\acro{relu}[RELU]{Rectified Linear Units}
\acro{rnn}[RNN]{Recurrent Neural Network}
\acro{rouge}[ROUGE]{Recall-Oriented Understudy for Gisting Evaluation}
\acro{sgd}[SGD]{Stochastic Gradient Descent}
\acro{sla}[SLA]{Sign Language Assessment}
\acro{slr}[SLR]{Sign Language Recognition}
\acro{slrt}[SLRT]{Sign Language Recognition Transformer}
\acro{slt}[SLT]{Sign Language Translation}
\acro{sltt}[SLTT]{Sign Language Translation Transformer}
\acro{sift}[SIFT]{Scale Invariant Feature Transform}
\acro{surf}[SURF]{Speeded Up Robust Features}
\acro{wer}[WER]{Word Error Rate}

\end{acronym}

\begin{document}
\pagestyle{headings}
\mainmatter
\def\ECCVSubNumber{12}  

\title{Multi-channel Transformers for\\Multi-articulatory Sign Language Translation} 

\titlerunning{Multi-channel Transformers for Multi-articulatory SLT}
%
\author{
Necati Cihan Camgoz\inst{1} \and
Oscar Koller\inst{2} \and
Simon Hadfield\inst{1} \and
Richard Bowden\inst{1}
}
\authorrunning{N. C. Camgoz et al.}
%
\institute{CVSSP, University of Surrey, UK, \{n.camgoz, s.hadfield, r.bowden\}@surrey.ac.uk,
\and
Microsoft, Munich, Germany, oscar.koller@microsoft.com
}

\maketitle

\begin{abstract}

Sign languages use multiple asynchronous information channels (articulators), not just the hands but also the face and body, which computational approaches often ignore. In this paper we tackle the multi-articulatory sign language translation task and propose a novel multi-channel transformer architecture. The proposed architecture allows both the inter and intra contextual relationships between different sign articulators to be modelled within the transformer network itself, while also maintaining channel specific information. We evaluate our approach on the RWTH-PHOENIX-Weather-2014T dataset and report competitive translation performance. Importantly, we overcome the reliance on gloss annotations which underpin other state-of-the-art approaches, thereby removing the need for expensive curated datasets.


\end{abstract}
\section{Introduction}

Sign languages are the main medium of communication of the Deaf. Every country typically has its own sign language and although some grammatical structures are shared, as are signs that rely upon heavy iconicity, different sign language have unique vocabularies \cite{stokoe1980sign,sutton1999linguistics}. Contrary to spoken and written languages, sign languages are visual. This makes automatic sign language understanding a novel research field where computer vision and natural language processing meet with a view to understanding and translating the spatio-temporal linguistic constructs of sign \cite{bragg2019sign}. 

Signers use multiple channels to convey information \cite{valli2000linguistics}. These channels, also known as articulators in linguistics \cite{malaia2018information}, can be grouped under two main categories with respect to their role in conveying information, namely manual and non-manual features \cite{brentari2018sign}. Manual features include the hand shape and its motion. Although manual features can be considered as the dominant part of the sign morphology, they alone do not encapsulate the full context of the conveyed information.
To give clarity, emphasis and additional meaning, signers use non-manual features, such as facial expressions, mouth gestures, mouthings\footnote{Mouthings are lip patterns that accompany a sign.} and body pose. Furthermore, both manual and non-manual features effect each other's meaning when used together.

To date, the literature in the field has predominantly focused on using the manual features to realize sign language recognition and translation \cite{parton2005sign,koller2016deephand,camgoz2018neural}, thus ignoring the rich and essential information contained in the non-manual features. This focus on the manual features is partially responsible for the common misconception that sign language recognition and translation problems are special sub-tasks of the gesture recognition field~\cite{pigou2014sign}. Sign language is as rich and complex as any spoken language. However, the multi-channel nature adds additional complexity as channels are not synchronised.

\begin{figure}[!t]
\centering
\includegraphics[trim={0cm 0cm 0cm 0.0cm},clip,width=0.95\linewidth]{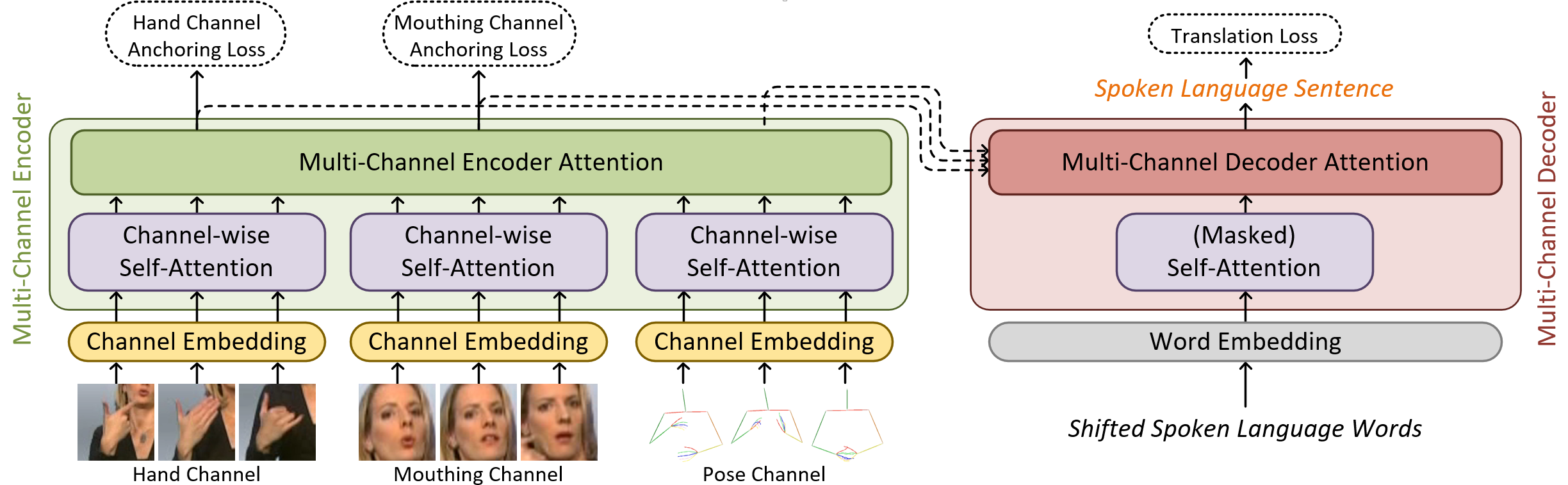}
\caption{An overview of the proposed Multi-channel Transformer architecture applied to the multi-articulatory \ac{slt} task.}
\label{fig:overview}
\end{figure}

In contrast to much of the existing literature, in this paper we model sign language by incorporating both manual and non-manual features into \ac{slt}. To achieve this, we utilize multiple channels which correspond to the articulatory subunits of the sign, namely hand shape, upper body pose and mouthings. We explore several approaches to combine the information present in these channels using both early and late fusion in a transformer architecture. Based on these findings we then introduce a novel deep learning architecture, the Multi-channel Transformer. This approach incorporates both inter and intra channel contextual relationships to learn meaningful spatio-temporal representations of asynchronous sign articulators, while also preserving channel specific information by using anchoring losses. Although this approach was designed specifically for \ac{slt}, we believe it can also be used to tackle other multi-channel sequence-to-sequence learning tasks, such as audio-visual speech recognition \cite{afouras2018deep}. An overview of the Multi-channel Transformer in the context of \ac{slt} can be seen in Figure \ref{fig:overview}. 

We evaluate our approach on the challenging \ac{ph14t} dataset which provides both sign gloss\footnote{Glosses can be considered as the minimal lexical items of the sign languages.} annotations and spoken language translations. Previous approaches \cite{camgoz2018neural,camgoz2020sign} on \ac{ph14t} heavily relied upon sign gloss annotations, which are labor intensive to obtain. We aim to remove this dependency on gloss annotation, by utilizing channel specific features obtained from related tasks, such as human pose estimation approaches \cite{cao2019openpose,Hidalgo_2019_ICCV} to represent upper body pose channel or lip reading features \cite{chung2017lip,assael2017lipnet} to represent mouthings \cite{koller2015deep,koller2014read}. Removing the dependency on manual annotation allows our approach to be scaled beyond what is possible with previous techniques, potentially using huge collections of un-annotated data. We empirically show that by integrating multiple articulator channels into our multi-channel transformer, it is possible to achieve competitive \ac{slt} performance which is \emph{on par} with models trained using additional gloss annotation.

The contributions of the paper can be summarized as: (1) We overcome the need for expensive gloss-level supervision by combining multiple articulatory channels with anchoring losses to achieve competitive continuous \ac{slt} performance on the \ac{ph14t} dataset. (2) We propose a novel multi-channel transformer architecture that supports multi-channel, asynchronous, sequence-to-sequence learning and (3) We use this to introduce the first successful approach to multi-articulatory \ac{slt}, which models the inter and intra relationship of manual and non-manual channels.

\section{Literature Review}
\label{sec:lit}

Computational processing of sign languages is an important field and expected to have tremendous impact on language deprivation of Deaf children, accessibility, sign linguistics and human-computer interaction in general. Its first attempt  dates back more than thirty years: A patent describing a hardwired electronic glove that recognized \ac{asl} finger spelling from hand configurations~\cite{grimes1983digital}. In the early days the field moved slowly, focusing first on isolated~\cite{tamura1988recognition}, then continuous sign language recognition~\cite{starner1998real}. With the rise of deep learning, enthusiasm was revived and accelerated the field~\cite{camgoz2016using,koller2016deephand,koller2016deepsign}. The recognition of limited domain but continuous real-life sign language became feasible~\cite{camgoz2017subunets,cui2017recurrent,huang2018video,koller2017resign,cui2019deep}. Driven by linguistic evidence~\cite{bellugi1972comparison,wilbur2000phonological,pfau2010nonmanuals}, the field realized that sign language recognition needs to focus on more than just the hands. Earlier works looked at several modalities separately, such as the face in general~\cite{vogler2008facial,koller2015continuous}, head pose~\cite{luzardo2013head}, the mouth~\cite{antonakos2012unsupervised,koller2014read,koller2015deep}, eye-gaze~\cite{caridakis2014nonmanual}, and body pose~\cite{pfister2012automatic,charles2014automatic}. More recently, multi-stream architectures showed strong performance~\cite{koller2019weakly,zhou2020spatialtemporal}.

Nevertheless, sign recognition only addresses part of the communication barrier between Deaf and hearing people. Sign languages follow a distinct grammar and are not word by word translations of spoken languages. After successful recognition, reordering and mapping into the target spoken language complete the communication pipeline.
In early works, recognition and translation were treated as two independent processing steps. Isolated single signs were recognized and subsequently translated~\cite{chai2013sign}. Often, existing work exclusively considered the problem as a text-to-text translation problem~\cite{bungeroth2004statistical,stein2010sign}, despite the visual nature of sign language and the lack of a written representation.

Generally speaking, much of the available sign translation literature falsely declares sign recognition as sign translation~\cite{fang2017deepasl,wang2018connectionist,guo2018hierarchical,guo2019dense}. Camgoz~\etal~\cite{camgoz2018neural} were the first to release a joint recognition and translation corpus with videos, glosses and translations to spoken language, covering real-life sign language, recorded from the broadcast news. They proposed to tackle the task based on a \ac{nmt} framework relying on input tokenization of the videos and subsequent sequence-to-sequence networks with attention. Their best performing tokenization method was based on strong sign recognition models trained using gloss annotations with full video frames and achieved an 18.1 BLEU-4 score, while a simple tokenization scheme (not trained with glosses) only reached 9.6 BLEU-4 on the test set of \ac{ph14t}.
Orbay and Akarun~\cite{orbay2020neural} investigated different tokenization methods on the same corpus and showed again that a pretrained hand shape recognizer~\cite{koller2016deephand} outperforms simpler approaches and reaches 14.6 BLEU-4. While they also investigated transformer architectures and multiple hands as input, the results underperformed. Ko~\etal~\cite{ko2019neural} describe a non-public dataset covering sign language videos, gloss annotation and translation. Their method relies on detected body keypoints only. It hence misses the important appearance based characteristics of sign. More recently, Camgoz~\etal~\cite{camgoz2020sign} proposed Sign Language Transformers, a joint end-to-end sign language recognition and translation approach. They used pre-trained gloss representations as inputs to their networks and trained transformer encoders using gloss annotations to learn meaningful spatio-temporal representations for \ac{slt}. Their approach is the current state-of-the-art on \ac{ph14t}. They report 20.2 BLEU-4 for pre-trained gloss features to spoken language translation, and 21.3 BLEU-4 with the additional gloss recognition supervision. 

Overall, previous work in the space of \ac{slt} has two major short-comings, which we intend to address with this paper: (1) The beauty of translations is the abundance of available training data, as they can be created in real-time by interpreters. Glosses are expensive to create and limit data availability. No previous work was able to achieve competitive performance while not relying on glosses. (2) So far \ac{slt} has never considered multiple articulators.

\section{Background on Neural Machine Translation}
\label{sec:background}

The objective of machine translation is to learn the conditional probability $p(\mathcal{Y}|\mathcal{X})$ where $\mathcal{X}=(x_1, ..., x_T)$ is a sentence from the source language with $T$ tokens and $\mathcal{Y}=(y_1, ..., y_U)$ is the desired corresponding translation of said sentence in the target language. To learn this mapping using neural networks, Kalchbrenner \etal \cite{kalchbrenner2013recurrent} proposed using an \emph{encoder-decoder} architecture, where the source sentence is encoded into a fixed sized ``context'' vector which is then used to decode the target sentence. Cho \etal \cite{cho2014learning} and Sutzeker \etal \cite{sutskever2014sequence} further improved this approach by assigning the encoding and decoding stages of translation to individual specialized \acp{rnn}.

The main drawback of \ac{rnn}-based approaches are long term dependency issues. Although there have been practical solutions to this, such as source sentence reversing  \cite{sutskever2014sequence}, the context vector is still of fixed size, and thus cannot perfectly encode arbitrarily long input sequences. To overcome the information bottleneck imposed by using the last hidden state of the \ac{rnn} as the context vector, Bahdanau \etal \cite{bahdanau2015neural} proposed an attention mechanism, which was a breakthrough in the field of \ac{nmt}. The idea behind the attention mechanism is to use a soft-search over the encoder outputs at each step of target sentence decoding. This was realized by conditioning target word prediction on a context vector which is a weighted sum of the source sentence representations. The weighting in turn is done by a learnt scoring function which measures the relevance of the decoders current hidden state and the encoder outputs. Luong \etal \cite{luong2015effective} further improved this approach by proposing a dot product attention (scoring) function as:\looseness=-1
\begin{equation}
    \mathrm{context} = \mathrm{softmax}\left(Q K^T\right)V 
\end{equation}
where queries, $Q$, correspond to the hidden state of the decoder at a given time step, and keys, $K$, and values, $V$, represent the encoder outputs.

More recently, Vaswani \etal \cite{vaswani2017attention} introduced self-attention mechanisms, which refine the source and target token representations by looking at the context they have been used in. Combining encoder and decoder self-attention layers with encoder-decoder attention, Vaswani \etal proposed Transformer networks, a fully connected network (as opposed to being \ac{rnn}-based) which has revolutionized the field of machine translation. In contrast to \ac{rnn}-based models, transformers obtain $Q$, $K$ and $V$ values by using individually learnt linear projection matrices at each attention layer. Vaswani \etal also introduced ``scaled'' dot-product attention as:
\begin{equation}
    \mathrm{context} = \mathrm{softmax}\left(\frac{Q K^T}{\sqrt{d_m}}\right) V 
\end{equation}
where $d_m$ is the number of hidden units of the model. The motivation behind the scaling operation is to counteract the effect of gradients becoming extremely small in cases where the number of hidden units is high and in-turn, the dot products grow large \cite{vaswani2017attention}. 

In this work we extend the transformer network architecture and adapt it to the task of multi-channel sequence-to-sequence learning. We propose a multi-channel attention layer to refine the representations of each source channel in the context of other source channels, while maintaining channel specific information using anchoring losses. We also adapt the encoder-decoder attention layer to be able to use multiple source channel representations. 

\begin{figure}[!t]
\begin{center}
\includegraphics[trim={0cm 0cm 0cm 0.0cm},clip,width=0.96\linewidth]{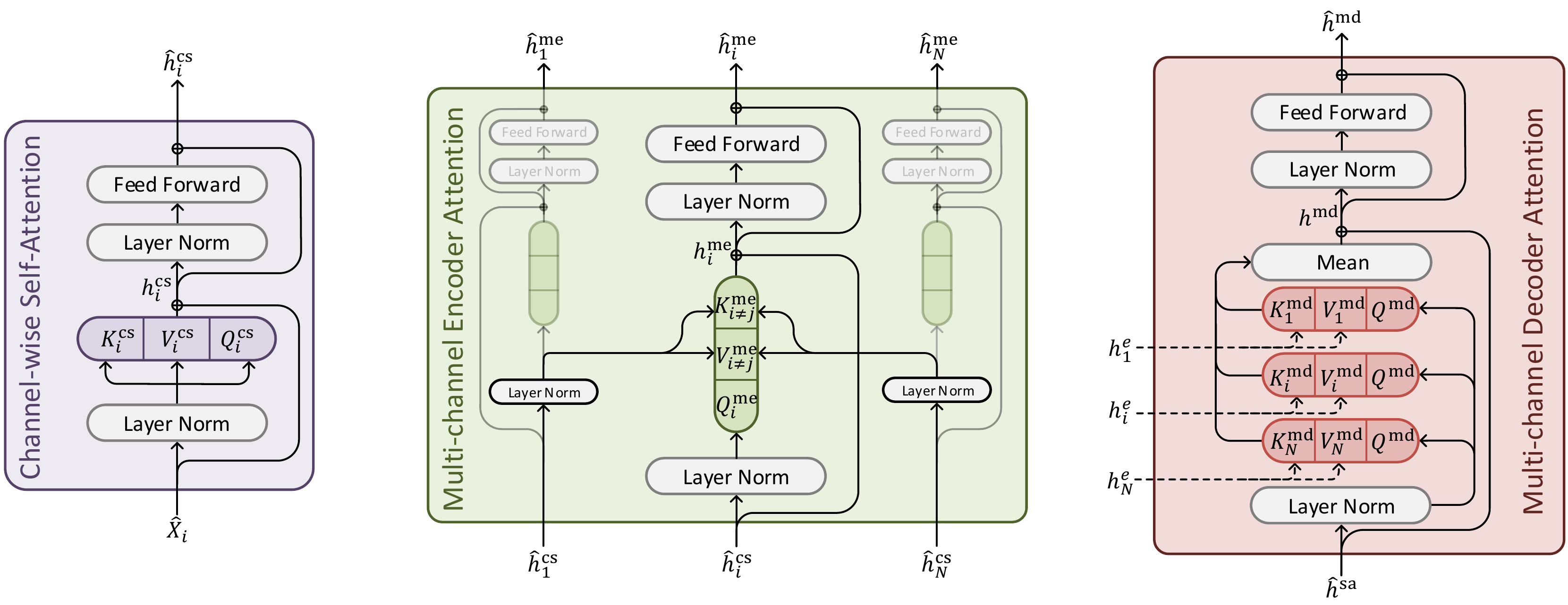}
\end{center}
\caption{A detailed overview of the introduced attention modules: (left) Channel-wise Self-Attention, (middle) Multi-channel Encoder Attention, and (right) Mult-channel Decoder Attention}
\label{fig:methodology}
\end{figure}

\section{Multi-channel Transformers}
\label{sec:method}

In this section we introduce Multi-channel Transformers, a novel architecture for sequence-to-sequence learning problems where the source information is embedded across several asynchronous channels. Given source sequences $\mathcal{X}=(X_{1}, ..., X_{N})$, where $X_i$ is the $i^{th}$ source channel with a cardinality of $T_{i}$, our objective is to learn the conditional probability $p(\mathcal{Y}|\mathcal{X})$, where $\mathcal{Y}=(y_1, ..., y_U)$ is the target sequence with $U$ tokens. In the application domain of \ac{slt}, these channels correspond to representations of the manual and non-manual features of the sign. An overview of the multi-channel transformer can be seen in Figure~\ref{fig:overview}, while individual attention modules introduced in this paper are visualized in Figure~\ref{fig:methodology}. To keep the formulation simple, and to focus the readers attention on the differentiating factors of our architecture, we omit the multi-headed attention, layer normalization and residual connections from our equations, which are the same as the original transformer networks \cite{vaswani2017attention}. 

\subsection{Channel and Word Embeddings}
As with other machine translation tasks, we start by projecting both the source channel features and the one-hot word vectors into a denser embedding where similar inputs lie close to one-another. To achieve this we use linear layers. We employ normalization and activation layers to change the scale of the embedded channel features and give additional non-linear representational capability to the model. The transformer networks do not have an implicit structure to model the position of a token within the sequence. To overcome this, we employ positional encoding \cite{vaswani2017attention} to add temporal ordering to the embedded representations. The embedding process for an input feature $x_{i,t}$ coming from the $i^{th}$ channel at time $t$ can be formalized as:\looseness=-1
\begin{align}
    \hat{x}_{i,t} &= \mathrm{Activ}\left(\mathrm{Norm}\left(x_{i,t} W^{\mathrm{ce}}_i + b^{\mathrm{ce}}_i\right)\right) + \mathrm{PosEnc}(t)
\end{align}
where $W^{\mathrm{ce}}_i$ and $b^{\mathrm{ce}}_i$ are channel specific learnt parameters of the linear projection layers. Similarly, the word embedding is as follows:
\begin{align}
    \hat{y}_{u}   &= y_u W^{\mathrm{we}} + b^{\mathrm{we}} + \mathrm{PosEnc}(u)    
\end{align}
where $W^{\mathrm{we}}$ and $b^{\mathrm{we}}$ are the weights of a linear layer which are either learned from scratch or pretrained on a large corpus \cite{bojanowski2017enriching,joulin2017bag}.

\subsection{Multi-Channel Encoder Layer}

\textbf{Channel-wise Self Attention (cs):} Each multi-channel encoder layer starts by learning the contextual relationships within a single channel by utilizing individual self-attention layers (See Figure~\ref{fig:methodology} (left)). As per the original transformer implementation, we use the scaled dot product scoring function in the attention mechanisms. Given embedded source channel representations, $\hat{X}_{i}$, we obtain $Q$ueries, $K$eys and $V$alues for the channel $i$ as\footnote{Note that we use a vectorized formulation in our equations. All softmax and bias addition operations are done row-wise.}:
\begin{align}
    \begin{split}
        Q^{\mathrm{cs}}_i &= \hat{X}_{i}W^{\mathrm{cs,q}}_{i} + b^{\mathrm{cs,q}}_{i} \\
        K^{\mathrm{cs}}_i &= \hat{X}_{i}W^{\mathrm{cs,k}}_{i} + b^{\mathrm{cs,k}}_{i} \\
        V^{\mathrm{cs}}_i &= \hat{X}_{i}W^{\mathrm{cs,v}}_{i} + b^{\mathrm{cs,v}}_{i} \\
    \end{split}
\end{align}
which are then passed to the channel-wise self attention function to have their intra channel contextual relationship modeled as:
\begin{equation}
    h^{\mathrm{cs}}_i = \mathrm{softmax}\left(\frac{Q^{\mathrm{cs}}_{i}\left(K^{\mathrm{cs}}_{i}\right)^{T}}{\sqrt{d_m}}\right) V^{\mathrm{cs}}_{i}
\end{equation}
where $h^{\mathrm{CS}}_i$ is the spatio-temporal representation of the $i^{th}$ source channel and $d_m$ is the hidden size of the model. We also utilize individual feed forward layers as described in \cite{vaswani2017attention} for each channel as:
\begin{equation}
    \mathrm{FF}(x) = \mathrm{max}\left(0, xW^{\mathrm{ff}}_1 + b_1\right) W^{\mathrm{ff}}_2 + b_2
\end{equation}
By feeding the contextually modeled channel representations through feed forward layers, we obtain the final outputs of the channel-wise attention layer of our multi-channel encoder layer as:
\begin{equation}
    \hat{h}^{\mathrm{cs}}_i = \mathrm{FF}^{\mathrm{cs}}_i(h^{\mathrm{cs}}_i)
\end{equation}

\noindent{}\textbf{Multi-channel Encoder Attention (me):} We now introduce the multi-channel encoder attention, which learns the contextual relationship between the self-attended channel representations (See Figure~\ref{fig:methodology} (middle)). As we are using dot product attention, we start by obtaining $Q$, $K$ and $V$ for each source as:
\begin{align}
    \begin{split}
        Q^{\mathrm{me}}_i &= \hat{h}^{\mathrm{cs}}_i W^{\mathrm{me,q}}_{i} + b^{\mathrm{me,q}}_{i} \\
        K^{\mathrm{me}}_i &= \hat{h}^{\mathrm{cs}}_i W^{\mathrm{me,k}}_{i} + b^{\mathrm{me,k}}_{i} \\
        V^{\mathrm{me}}_i &= \hat{h}^{\mathrm{cs}}_i W^{\mathrm{me,v}}_{i} + b^{\mathrm{me,v}}_{i} \\
    \end{split}
\end{align}
These values are then passed to the multi-channel attention layers where the $Q$ueries of each channel are used to estimate the scores over the concatenated $K$eys of the other channels. These scores are then used to calculate the channel-fused representations by taking a weighted sum over the other channels' concatenated $V$alues. More formally, multi-channel attention can be defined as:
\begin{equation}
    h^{\mathrm{me}}_i = \mathrm{softmax} \left(\frac{Q^{\mathrm{me}}_{i}\left(\left[\forall K^{\mathrm{me}}_{j} \; \mathrm{where} \; j \neq i \right]\right) ^{T}}{\sqrt{d_m}}\right) \left[\forall V^{\mathrm{me}}_{j} \; \mathrm{where} \; j \neq i \right]
\end{equation}
We would like to note that, the concatenation operation ($[\;\;]$) is performed over the time axis, thus making our approach applicable to tasks where the source channels have a different numbers of tokens. We then pass multi-channel attention outputs to individual feed forward layers to obtain the final outputs of the multi-channel encoder layer as:
\begin{equation}
    \hat{h}^{\mathrm{me}}_i = \mathrm{FF}^{\mathrm{me}}_i(h^{\mathrm{me}}_i)
\end{equation}

Several multi-channel encoder layers can be stacked to form the encoder network with the aim of learning more complex multi-channel contextual representations, ${h^{e} = (h^{e}_{1}, ..., h^{e}_{N})}$, where $h^e_i$ is the output corresponding to the $i^{th}$ source channel.

\subsection{Multi-channel Decoder Layer}
Transformer networks utilize a masked self attention and an encoder-decoder attention in each decoder layer. The subsequent masking on self-attention is essential, as the target tokens' successors will not be available at inference time. In our approach, we also employ the masked self-attention to model the contextual relationship between target tokens' and its predecessors. However, we replace encoder-decoder attention with multi-channel decoder attention, which is modified to work with multiple source channel representations (See Figure~\ref{fig:methodology} (right)). Given the word embeddings $\hat{\mathcal{Y}}$ of a sentence $\mathcal{Y}$, we first obtain the masked self-attention (sa) outputs $\hat{h}^\mathrm{sa}$ using the generic approach \cite{vaswani2017attention}, which are then in turn passed to our multi-channel decoder attention.
 
\noindent{}\textbf{Multi-channel Decoder Attention (md):} In generic transformers, encoder-decoder attention $Q$ueries are obtained from the decoder self-attention estimates, $\hat{h}^\mathrm{sa}$, while $K$eys and $V$alues are calculated from the final encoder layer outputs, $h_e$. In order to incorporate information coming from multiple channels using transformer models, we propose the multi-channel decoder attention module. We first obtain the $Q$, $K$ and $V$ as:
\begin{align}
    \begin{split}
        Q^{\mathrm{md}} &= \hat{h}^\mathrm{sa} W^{\mathrm{md,q}} + b^{\mathrm{md,q}} \\
        K^{\mathrm{md}}_i &= h^{e}_i W^{\mathrm{md,k}}_{i} + b^{\mathrm{md,k}}_{i} \\
        V^{\mathrm{md}}_i &= h^{e}_i W^{\mathrm{md,v}}_{i} + b^{\mathrm{md,v}}_{i} \\
    \end{split}
\end{align}
Note that each source channel $i$ has their own learned $K$ey and $V$alue matrices, $W_i^{md,k}$ and $W_i^{md,v}$ respectively. 

These are then passed to the multi-channel decoder attention module where the $Q$ueries of each target token are scored against all channel $K$eys. Channel scores are then used to calculate the weighted average of their respective $V$alues. Individual channel outputs are averaged to obtain the final output of the multi-channel decoder attention module.  This process can be formalized as:
\begin{equation}
    h^{\mathrm{md}} = \frac{1}{N} \sum_{i=1}^{N} \left(\mathrm{softmax} \left(\frac{Q^{\mathrm{md}}\left(K^{\mathrm{md}}_{i}\right)^T}{\sqrt{d_m}}\right) V^{\mathrm{md}}_i\right)
\end{equation}

The attention module outputs are then passed through a feed forward layer to obtain the final representations of the multi-channel decoder layer as:
\begin{equation}
    \hat{h}^{\mathrm{md}} = \mathrm{FF}^{\mathrm{md}}(h^{\mathrm{md}})
\end{equation}

Like the multi-channel encoder layer, multiple decoder layers can be stacked to improve representation capabilities of the decoder network. The output of the stacked decoder is denoted as $h^d = (h_1, ..., h_U)$ which is used to condition target token generation.

\subsection{Loss Functions}

We propose training multi-channel transformers using two types of loss function, namely a Translation Loss, which is commonly used in machine translation, and a Channel Anchoring Loss, which aims to preserve channel specific information during encoding.

\textbf{Translation Loss:} Although different loss functions have been used to train translation models, such as a mixture-of-softmaxes \cite{yang2018breaking}, token level cross-entropy loss is the most common approach to learn network parameters. Given a source-target pair, the translation loss, $\mathcal{L}_T$, is calculated as the accumulation of the error at each decoding step $u$, which is estimated using a classification loss over the target vocabulary as:
\begin{equation}
\label{multitrans:eq:loss}
    \mathcal{L}_T = 1 - \prod_{u=1}^{U} \sum_{g=1}^G p(y^g_u) p(\hat{y}^g_u) 
\end{equation}
where $p(y^g_u)$ and $p(\hat{y}^g_u)$ represent the ground truth and the generation probabilities of the target $y^g$ at decoding step $u$, respectively, and $G$ is the target language vocabulary size. In our networks, the probability of generating target token $y_u$ at the decoding step $u$ is conditioned on the hidden state of the decoder network $h^d_u$ at the corresponding time step, $p(\hat{y}_u) = p(\hat{y}_u|h^d_u)$. Softmaxed linear projection of $h^d_u$ is used to model the probability of producing tokens over the whole target vocabulary as:
\begin{equation}
    p(\hat{y}_u|h^d_u) = \mathrm{softmax}(h^d_u W^{o} + b^{o})
\end{equation}
where $W^{o}$ and $b^{o}$ are the trainable parameters of a linear layer. 

\textbf{Channel Anchoring Loss:} For source channels, where we have access to a relevant classifier, we use an anchoring loss to preserve channel specific information. Predictions of these classifiers are used as ground truth to calculate token level cross entropy losses in the same manner as the translation loss. Given the classifier outputs corresponding to the $i^{th}$ channel, ${C_i = (c_{i,1}, ..., c_{i,T_{i}})}$, and the hidden state of the encoder, $h^e$, we first calculate the prediction probabilities over the target channel classes as:
\begin{equation}
    p(\hat{c}_{i,t}|h^e) = \mathrm{softmax}(h^e W_i^{o} + b_i^{o})
\end{equation}
where $p(\hat{c}_{i,t}|h^e)$ represent the prediction probabilities over the $i^{th}$ channel's classifier vocabulary, while $W_i^{o}$ and $b_i^{o}$ are the weights and biases of the linear layer used for the $i^{th}$ channel, respectively. We then use a modified version of Equation~\ref{multitrans:eq:loss} to calculate the $i^{th}$ channel's anchoring loss, $\mathcal{L}_{A,i}$, as:
\begin{equation}
\label{multitrans:eq:loss:anchor}
    \mathcal{L}_{A,i} = 1 - \prod_{t=1}^{T} \sum_{g=1}^{G_i} p(c^g_{i,t}) p(\hat{c}^g_{i,t}|h^e_t) 
\end{equation}
where $p(c^g_{i,t})$ and $p(\hat{c}^g_{i,t})$ represent the classifier output and the predicted probabilities of the class $c^d_i$ at the encoders $t^{th}$ step, respectively, while $G_i$ is the number of target classes of the classifier corresponding to channel $i$. For example, one can use a hand shape classifier's convolutional layer as as input channel and the same classifier's predictions as ground truth for hand channel anchoring loss to preserve the hand shape information, as well as to regularize the translation loss. 

\textbf{Total Loss:} We use a weighed combination of Translation loss, $\mathcal{L}_{T}$, and Anchoring losses, $\mathcal{L}_{A} = (\mathcal{L}_{A,1}, ..., \mathcal{L}_{A,N})$, during training as:
\begin{equation}
    \mathcal{L} = \lambda_{T}\mathcal{L}_{T} + \lambda_{A} \sum_{i=1}^{N} \mathcal{L}_{A,i}
\end{equation}
where $\lambda_{T}$ and $\lambda_{A}$ decide the importance of each loss function during training. 

\section{Implementation and Evaluation Details}
\label{sec:imp}

\hspace{\parindent}\textbf{Dataset:} We evaluate our model on the challenging \ac{ph14t} \cite{camgoz2018neural} dataset, which is currently the only publicly available large vocabulary continuous \ac{slt} dataset aimed at vision based sign language research. 

\textbf{Sign Channels:} We use three different articulators/channels to represent the manual and non-manual features of the sign, namely hand shapes, mouthings and upper body pose. We employ the models proposed and used in \cite{koller2019weakly} and from these networks extracted 1024 dimensional \ac{cnn} features for each frame (last layer before the fully connected layer) for hand shape and mouthing channels. We use the class prediction from the same network to anchor the channel representations. Although these networks were trained on 61 and 40 hand shapes and mouthing classes respectively (including transition/silence class), the predictions only contained 52 and 36 classes. Hence, our anchoring losses are calculated over the predicted number of classes.

To represent the upper body pose of the signers, we extract 2D skeletal pose information using the OpenPose library \cite{cao2019openpose}. We then employ a 2D-to-3D lifting approach designed specifically for sign language production to obtain the final 3D joint positions of 50 upper body pose joints \cite{zelinka2019nn}. As there were no prior subunit classes for the upper body pose on \ac{ph14t}, we do not utilize an anchoring loss on the pose channel. 

\textbf{Training and Network Details:} Our networks are trained using the PyTorch framework \cite{paszke2017automatic} with a modified version of the JoeyNMT library \cite{kreutzer2019joey}. We use Adam \cite{kingma2014adam} optimizer with a batch size of $32$, a learning rate of $10^{-3}$ ($\beta_1=0.9, \beta_2={0.998}$) and a weight decay of $10^{-3}$. We utilize Xavier \cite{glorot2010understanding} initialization and train all networks from scratch. We do not apply dropout and only use a single headed scaled dot-product attention to reduce the number of hyper-parameters in our experiments.

\textbf{Decoding:} During training we use a greedy search to evaluate development set translation performance. At inference, we employ beam search decoding with the beam width ranging from 0 to 10. We also employ a length penalty as proposed by \cite{johnson2017google} with $\alpha$ values ranging from $0$ to $5$. We use the development set to find the best performing beam width and $\alpha$, and use these during test set inference for final results.

\textbf{Performance Metrics:} We use BLEU \cite{papineni2002bleu} and ROUGE \cite{lin2004rouge} scores to measure the translation performance. To give the reader a better understanding of the networks behaviour, we repeat each experiment 10 times and report mean and standard deviation of BLEU-4 and ROUGE scores on both development and test sets. We also report our best result for every setup based on the BLEU-4 score as per the development set. BLEU-4 score is also used as the validation score for our learning scheduler and for early stopping. 

\section{Experiment Results}
\label{sec:quant}

In this section we propose several multi-channel \ac{slt} experimental setups and report our quantitative results. We start by sharing single channel \ac{slt} performance using different network architectures, varying the number of hidden units, both to set a baseline for our multi-channel approaches and to find the optimal network size. After that, we propose two naive channel fusion approaches, namely early fusion and late fusion, to set a fusion benchmark for our novel \emph{Multi-channel Transformer} architecture. Finally, we report the performance of the multi-channel transformer approach with and without the channel anchoring losses and compare our results against the state-of-the-art.  

We apply batch normalization \cite{ioffe2015batch} and a soft-sign activation function \cite{nwankpa2018activation} to input channel embeddings before passing them to our networks. See supplementary material for empirical justification for this choice.

\subsection{Single Channel Baselines}
In the first set of experiments, we train single channel \ac{slt} models. The main objective of these experiments is to set translation baselines for all future multi-channel fusion models. However, we would also like to examine the relative information presented in each channel by comparing their translation performance against one another. In addition, we wish to identify the optimal network setup for each channel to guide the future experiments. Therefore, we conduct experiments with four network setups for all three articulators with sizes varying from 32x64 to 256x512 (hidden size (HS) x number of feed forward (FF) units). All networks were built using two encoder and decoder layers.

As can be seen in Table~\ref{tbl:sng:chn}, \textbf{H}and is the best performing channel in all network setups. Furthermore, using a network setup of 128x256 outperforms all of the alternatives. We believe this is closely related to the limited number of training samples we have and the over-fitting issues that come with it. Therefore, for the rest of our experiments we use 128x256 parameters for each channel. 

\begin{table}[!t]
\centering
\caption{Single channel \ac{slt} baselines using different network architectures.}
\label{tbl:sng:chn}
\resizebox{0.80\linewidth}{!}{
\begin{tabular}{ll|cc|cc|cc|cc}
&  &\multicolumn{4}{c|}{Dev Set} & \multicolumn{4}{c}{Test Set} \\
 &  & \multicolumn{2}{c|}{BLEU-4} & \multicolumn{2}{c|}{ROUGE} & \multicolumn{2}{c|}{BLEU-4} & \multicolumn{2}{c}{ROUGE} \\ \hline
 Channel & HSxFF & Best  & mean $\pm$ std   & -     & mean $\pm$ std  & -     & mean $\pm$ std   & -     & mean $\pm$ std   \\ \hline
 \hline
\textbf{H}and       & 32x64  & 14.54 & 14.05 $\pm$ 0.42 & 38.47 & 38.49 $\pm$ 0.45 & 13.88 & 13.80 $\pm$ 0.63 & 38.05 & 38.04 $\pm$ 0.49 \\ 
\textbf{H}and       & 64x128 & \textbf{16.44} & 15.70 $\pm$ 0.41 & 40.79 & 40.45 $\pm$ 0.64 & 16.18 & 15.63 $\pm$ 0.65 & 40.62 & 40.07 $\pm$ 0.80 \\ 
\textbf{H}and       & 128x256& 16.32 & \textbf{15.91} $\pm$ 0.43 & \textbf{41.87} & \textbf{41.08} $\pm$ 0.73 & \textbf{16.76} & \textbf{16.02} $\pm$ 0.88 & \textbf{41.85} & \textbf{40.67} $\pm$ 0.93 \\ 
\textbf{H}and       & 256x512& 16.06 & 15.41 $\pm$ 0.46 & 41.46 & 40.13 $\pm$ 0.83 & 15.43 & 15.57 $\pm$ 0.60 & 40.48 & 39.88 $\pm$ 0.71 \\ 
\hline
\textbf{M}outhing   & 32x64  & 11.70 & 11.24 $\pm$ 0.34 & 33.26 & 32.84 $\pm$ 0.60 & 10.77 & 11.01 $\pm$ 0.34 & 33.51 & 33.05 $\pm$ 0.34 \\ 
\textbf{M}outhing   & 64x128 & 12.91 & 12.55 $\pm$ 0.30 & 36.22 & 35.17 $\pm$ 0.71 & 12.83 & 12.62 $\pm$ 0.50 & 35.30 & 35.04 $\pm$ 0.72 \\ 
\textbf{M}outhing   & 128x256&\textbf{13.74} &\textbf{13.08} $\pm$ 0.41 &\textbf{37.20} &\textbf{35.96} $\pm$ 0.79 &\textbf{13.77} &\textbf{13.50} $\pm$ 0.37 &\textbf{37.24} &\textbf{36.60} $\pm$ 0.72 \\ 
\textbf{M}outhing   & 256x512& 12.86 & 12.40 $\pm$ 0.42 & 34.13 & 34.63 $\pm$ 0.64 & 13.25 & 12.34 $\pm$ 0.51 & 35.53 & 34.83 $\pm$ 0.47 \\ 
\hline
\textbf{P}ose       & 32x64  &  9.64 &  8.91 $\pm$ 0.42 & 30.27 & 29.92 $\pm$ 0.67 &  9.64 &  8.55 $\pm$ 0.62 & 30.03 & 29.13 $\pm$ 0.79 \\ 
\textbf{P}ose       & 64x128 & 10.64 & 10.28 $\pm$ 0.26 & 31.06 & 31.31 $\pm$ 0.47 &  9.97 &  9.88 $\pm$ 0.18 & 29.63 & 30.44 $\pm$ 0.60 \\ 
\textbf{P}ose       & 128x256&\textbf{11.02} &\textbf{10.52} $\pm$ 0.31 &\textbf{32.22} &\textbf{31.85} $\pm$ 0.42 &\textbf{10.26} &\textbf{10.03} $\pm$ 0.46 &\textbf{30.44} &\textbf{30.79} $\pm$ 0.82 \\ 
\textbf{P}ose       & 256x512& 10.06 &  9.50 $\pm$ 0.51 & 31.03 & 30.11 $\pm$ 0.75 &  9.51 &  8.62 $\pm$ 0.70 & 29.92 & 28.65 $\pm$ 0.99 \\ 
\hline
\hline
\textbf{G}loss      & 32x64  & 17.21 & 16.03 $\pm$ 0.49 & 42.20 & 41.26 $\pm$ 0.64 & 15.45 & 15.68 $\pm$ 0.43 & 41.21 & 40.71 $\pm$ 0.42 \\ 
\textbf{G}loss      & 64x128 & 18.50 & 18.16 $\pm$ 0.23 & 44.99 & 43.87 $\pm$ 0.68 & 18.14 & 17.89 $\pm$ 0.56 & 43.57 & 43.02 $\pm$ 0.69 \\ 
\textbf{G}loss      & 128x256& 19.43 &\textbf{19.14} $\pm$ 0.36 &\textbf{46.10} &\textbf{45.17} $\pm$ 0.63 & 19.52 &\textbf{19.08} $\pm$ 0.48 &\textbf{45.32} &\textbf{44.52} $\pm$ 0.80 \\ 
\textbf{G}loss      & 256x512&\textbf{19.52} & 18.36 $\pm$ 0.50 & 45.97 & 44.16 $\pm$ 0.79 &\textbf{19.61} & 18.60 $\pm$ 0.63 & 45.29 & 43.92 $\pm$ 0.98 \\ 
\hline
\end{tabular}
 }
\end{table}

We further train a \textbf{G}loss single channel network to set a baseline for our multi-channel approaches to compare against. As shown in Table~\ref{tbl:sng:chn}, using \ac{cnn} features that were trained using gloss level annotations outperforms all single sign articular based models (19.52 \textit{vs.} 16.44 dev BLEU-4 score). Although the 256x512 network setup obtained the best individual development and test set translation performances, the mean performance of the 128x256 network was better, encouraging us to utilize this setup going forward.

\subsection{Early and Late Fusion of Sign Channels}
To set another benchmark for our multi-channel transformer, we propose two naive multi-channel fusion approaches, namely early and late fusion. In the early fusion setup, features from different channels are concatenated to create a fused representation of each frame. These representations are then used to train \ac{slt} models, as if they were features coming from a single channel. Hence, the contextual relationship is performed in an implicit manner by the transformer architecture. In our second, late fusion setup, individual \ac{slt} models are built which are then fused at the decoder output level, \ie $h^d$, by concatenation. The fused representation is then used to generate target tokens using a linear projection layer. Compared to early fusion, this approach's capability to learn more abstract relationships is limited as the fusion is only done by a single linear layer. We examine all four possible fusions of the three channels. Network setup is set to linearly scale with respect to the number of channels that are fused together with a factor of 128x256 per channel.

\begin{table}[!t]
\caption{\ac{slt} performance of early and late channel fusion approaches.}
\label{tbl:early} 
\centering
\resizebox{0.85\linewidth}{!}{
\begin{tabular}{lll|cc|cc|cc|cc}
 & & &\multicolumn{4}{c|}{Dev Set} & \multicolumn{4}{c}{Test Set} \\
 & & & \multicolumn{2}{c|}{BLEU-4} & \multicolumn{2}{c|}{ROUGE} & \multicolumn{2}{c|}{BLEU-4} & \multicolumn{2}{c}{ROUGE} \\ \hline
Fusion & Channels & HSxFF & Best  & mean $\pm$ std   & -     & mean $\pm$ std  & -     & mean $\pm$ std   & -     & mean $\pm$ std   \\ \hline
Early & \textbf{H} + \textbf{M}              & 2*(128x256) & \textbf{17.25} & \textbf{16.73} $\pm$ 0.57 & \textbf{42.04} & \textbf{41.72} $\pm$ 0.61 & \textbf{17.37} & \textbf{16.73} $\pm$ 0.82 & \textbf{42.35} & \textbf{41.76} $\pm$ 0.80 \\ 
Early & \textbf{H} + \textbf{P}              & 2*(128x256) & 16.17 & 15.70 $\pm$ 0.32 & 40.51 & 40.28 $\pm$ 0.46 & 15.75 & 15.83 $\pm$ 0.30 & 40.54 & 40.34 $\pm$ 0.70 \\ 
Early & \textbf{M} + \textbf{P}              & 2*(128x256) & 13.57 & 12.91 $\pm$ 0.30 & 36.43 & 35.88 $\pm$ 0.40 & 13.23 & 13.02 $\pm$ 0.42 & 35.66 & 36.01 $\pm$ 0.72 \\ 
Early & \textbf{H} + \textbf{M} + \textbf{P} & 3*(128x256) & 15.69 & 15.08 $\pm$ 0.55 & 39.78 & 39.38 $\pm$ 0.73 & 15.19 & 15.19 $\pm$ 0.49 & 39.99 & 39.43 $\pm$ 0.80 \\ 
\hline 
\hline
Late & \textbf{H} + \textbf{M}              & 2*(128x256)  & \textbf{17.03} & \textbf{16.36} $\pm$ 0.48 & 41.69 & \textbf{41.58} $\pm$ 0.79 & 16.81 & \textbf{16.67} $\pm$ 0.49 & 41.69 & \textbf{41.69} $\pm$ 0.54 \\ 
Late & \textbf{H} + \textbf{P}              & 2*(128x256)  & 16.61 & 16.16 $\pm$ 0.33 & 41.54 & 41.18 $\pm$ 0.58 & 15.90 & 16.07 $\pm$ 0.76 & 40.81 & 40.50 $\pm$ 0.99 \\ 
Late & \textbf{M} + \textbf{P}              & 2*(128x256)  & 14.22 & 13.55 $\pm$ 0.31 & 36.44 & 37.03 $\pm$ 0.64 & 14.11 & 13.65 $\pm$ 0.49 & 36.25 & 36.95 $\pm$ 0.65 \\ 
Late & \textbf{H} + \textbf{M}  + \textbf{P} & 3*(128x256) & 17.00 & 16.35 $\pm$ 0.38 & \textbf{42.09} & 41.29 $\pm$ 0.42 & \textbf{16.95} & 16.50 $\pm$ 0.47 & \textbf{42.12} & 41.53 $\pm$ 0.52 \\ 
\hline
\end{tabular}}
\end{table}

As can be seen in Table~\ref{tbl:early}, fusion of \textbf{H}ands and \textbf{M}outh yields slightly better results than single channel translation models (excluding gloss). However, unlike late fusion, which saw improvement in all scenarios, early fusion's performance gets worse as more features are added to the network. As this means having more parameters in our networks, we believe this is due to the natural propensity of the transformers to over-fit on small training datasets, like ours. 

\subsection{Multi-channel Transformers}
In this set of experiments we examine the translation performance of the proposed multi-channel transformer architecture for multi-articulatory \ac{slt}. We first start by investigating the effects of the anchoring loss. We then compare our best performing method against other fusion options, gloss based translation and other state-of-the-art methods. As with other fusion experiments, we examine all possible fusion combinations. In addition to using the 128x256 network setup, we also evaluate having a larger network to see if the additional anchoring losses help with over-fitting by regularizing the translation loss.

As can be seen in the first row of Table~\ref{tbl:ours}, while using the same number of parameters as the early and late fusion setups, our proposed \emph{Multi-Channel Transformer} approach outperforms both configurations. However, doubling the network size does effect the direct application of multi-channel attention negatively. To counteract this issue and to examine the effects of the anchoring loss, we run experiments with both 128x256 and 256x512 setups. We normalize our losses on the sequence level instead of token level and we set the anchoring loss weight, $\lambda_{A}$, to 0.15 to counteract different source (video) and target (sentence) sequence lengths. Using the anchoring losses not only improves the performance of the 128x256 models but also allows the 256x512 networks to achieve similar translation performance to using gloss features. We believe this is due to two main factors. Firstly, the anchoring loss forces the encoder channels to preserve the channel specific information while being contextually modeled against other articulators. Secondly, it acts as a regularizer for the translation loss and counteracts the over-fitting previously discussed. 

\begin{table}[!t]
\caption{Multi-channel Transformer based multi-articulatory \ac{slt} results.}
\centering
\resizebox{0.95\linewidth}{!}{
\begin{tabular}{lcc|cc|cc|cc|cc}
 & & & \multicolumn{4}{c|}{Dev Set} & \multicolumn{4}{c}{Test Set} \\
 & & & \multicolumn{2}{c|}{BLEU-4} & \multicolumn{2}{c|}{ROUGE} & \multicolumn{2}{c|}{BLEU-4} & \multicolumn{2}{c}{ROUGE} \\ \hline
  Channels & Anchoring Loss & HSxFF & Best  & mean $\pm$ std   & -     & mean $\pm$ std  & -     & mean $\pm$ std   & -     & mean $\pm$ std   \\ \hline
\hline
\textbf{H} + \textbf{M}  & \crx & 2*(128x256)  & 17.71 & 16.97 $\pm$ 0.53 & 43.43 & 42.02 $\pm$ 0.86 & 17.72 & 17.19 $\pm$ 0.73 & 42.70 & 41.95 $\pm$ 0.85 \\ 
\textbf{H} + \textbf{P}  & \crx & 2*(128x256)  & 17.20 & 16.36 $\pm$ 0.58 & 42.15 & 41.23 $\pm$ 0.46 & 16.41 & 16.25 $\pm$ 0.66 & 40.56 & 40.87 $\pm$ 0.67 \\ 
\textbf{M} + \textbf{P}  & \crx & 2*(128x256)  & 14.17 & 13.50 $\pm$ 0.40 & 36.82 & 36.62 $\pm$ 0.52 & 13.43 & 13.93 $\pm$ 0.44 & 37.03 & 37.43 $\pm$ 0.65 \\ 
\textbf{H} + \textbf{M}  + \textbf{P} &  \crx & 3*(128x256) & 17.98 & 16.89 $\pm$ 0.59 & 44.01 & 41.85 $\pm$ 0.93 & 17.15 & 16.85 $\pm$ 0.65 & 42.38 & 41.83 $\pm$ 0.85 \\ 
\hline
\textbf{H} + \textbf{M}  &  \crx & 2*(256x512)  & 15.95 & 15.46 $\pm$ 0.34 & 41.01 & 40.31 $\pm$ 0.50 & 15.87 & 15.80 $\pm$ 0.36 & 40.30 & 40.40 $\pm$ 0.70 \\ 
\textbf{H} + \textbf{P}  &  \crx & 2*(256x512)  & 15.41 & 14.95 $\pm$ 0.33 & 40.10 & 39.22 $\pm$ 0.53 & 15.91 & 15.14 $\pm$ 0.59 & 40.24 & 39.11 $\pm$ 0.69 \\ 
\textbf{M} + \textbf{P}  &  \crx & 2*(256x512)  & 13.39 & 12.60 $\pm$ 0.49 & 35.70 & 35.01 $\pm$ 0.73 & 13.38 & 12.74 $\pm$ 0.48 & 36.89 & 35.39 $\pm$ 0.79 \\ 
\textbf{H} + \textbf{M}  + \textbf{P} & \crx & 3*(256x512) & 15.87 & 14.97 $\pm$ 0.51 & 40.53 & 39.65 $\pm$ 0.79 & 16.02 & 15.17 $\pm$ 0.81 & 40.15 & 39.61 $\pm$ 1.12 \\ 
\hline
\textbf{H} + \textbf{M}   & \cbt & 2*(128x256)  & 18.52 & 17.93 $\pm$ 0.39 & 44.56 & 43.25 $\pm$ 0.57 & 17.93 & 17.76 $\pm$ 0.49 & 43.21 & 42.91 $\pm$ 0.49 \\ 
\textbf{H} + \textbf{P}  & \cbt & 2*(128x256) & 17.70 & 16.53 $\pm$ 0.67 & 43.19 & 41.26 $\pm$ 0.99 & 16.93 & 16.41 $\pm$ 0.48 & 42.62 & 40.80 $\pm$ 0.82 \\ 
\textbf{M} + \textbf{P}  & \cbt & 2*(128x256)  & 15.14 & 14.60 $\pm$ 0.32 & 38.57 & 38.45 $\pm$ 0.45 & 15.32 & 15.05 $\pm$ 0.72 & 38.47 & 38.66 $\pm$ 0.76 \\ 
\textbf{H} + \textbf{M}  + \textbf{P} & \cbt & 3*(128x256) & 18.80 & 17.81 $\pm$ 0.68 & 44.24 & 43.17 $\pm$ 0.81 & 18.30 & 17.75 $\pm$ 0.58 & 43.65 & 42.90 $\pm$ 0.62 \\ 
\hline
\textbf{H} + \textbf{M}   & \cbt & 2*(256x512)  & 19.05 & 18.07 $\pm$ 0.44 & 45.04 & 43.76 $\pm$ 0.82 &\textbf{ 19.21} & 17.71 $\pm$ 0.72 & \textbf{45.05} & 43.29 $\pm$ 0.99 \\ 
\textbf{H} + \textbf{P}  & \cbt & 2*(256x512)  & 16.80 & 16.29 $\pm$ 0.36 & 41.15 & 40.86 $\pm$ 0.42 & 16.68 & 16.29 $\pm$ 0.48 & 41.34 & 40.68 $\pm$ 0.76 \\ 
\textbf{M} + \textbf{P}  & \cbt & 2*(256x512)  & 15.14 & 14.60 $\pm$ 0.26 & 39.45 & 38.47 $\pm$ 0.62 & 15.36 & 15.13 $\pm$ 0.44 & 40.06 & 39.09 $\pm$ 0.62 \\ 

\textbf{H} + \textbf{M}  + \textbf{P} & \cbt & \textbf{3*(256x512)} & \textbf{19.51} & \textbf{18.66} $\pm$ 0.52 & \textbf{45.90} & \textbf{44.30} $\pm$ 0.92 & 18.51 & \textbf{18.31} $\pm$ 0.57 & 43.57 & \textbf{43.75} $\pm$ 0.63 \\ 
\hline 
\textbf{G}loss & --  & 128x256& 19.43 & \textbf{19.14} $\pm$ 0.36 & \textbf{46.10} & \textbf{45.17} $\pm$ 0.63 & 19.52 & \textbf{19.08} $\pm$ 0.48 & \textbf{45.32} & \textbf{44.52} $\pm$ 0.80 \\ 
\textbf{G}loss &  -- & \textbf{256x512}&\textbf{19.52} & 18.36 $\pm$ 0.50 & 45.97 & 44.16 $\pm$ 0.79 &\textbf{19.61} & 18.60 $\pm$ 0.63 & 45.29 & 43.92 $\pm$ 0.98 \\ 
\hline 
Orbay \etal \cite{orbay2020neural} & -- & -- & -- & -- & -- & -- & 14.56 & -- & 38.05 & -- \\
Sign2Gloss$\rightarrow$Gloss2Text~\cite{camgoz2018neural} & -- & -- & 17.89 & -- & 43.76 & -- & 17.79 & -- & 43.45 & -- \\
Sign2Gloss2Text~\cite{camgoz2018neural} & -- & -- & 18.40 & -- & 44.14 & -- & 18.13 & -- & 43.80 & -- \\
(Gloss) Sign2Text~\cite{camgoz2020sign} & -- & \textbf{3x(512x2048)} & \textbf{20.69} & -- & -- & -- & \textbf{20.17} & -- & -- & -- \\
(Gloss) Sign2(Gloss+Text)~\cite{camgoz2020sign} & -- & \textbf{3x(512x2048)} & \textbf{22.38} & -- & -- & -- & \textbf{21.32} & -- & -- & -- \\
\hline 
\end{tabular}
}
\label{tbl:ours} 

\end{table}

Compared to the state-of-the-art, our best multi-channel transformer model surpasses the performance of several previous models \cite{orbay2020neural,camgoz2018neural}, some of which are heavily reliant on gloss annotation. Furthermore, our multi-channel models perform \textit{on par} with our single gloss channel model, and yields competitive translation performance to the state-of-the-art transformer based approaches \cite{camgoz2020sign}, which utilize larger models and uses gloss supervision on several levels (pretrained Gloss CNN features and transformer encoder supervision). However, due to their dependence on gloss annotations, such models \cite{camgoz2018neural,camgoz2020sign} can not be scaled to larger un-annotated datasets, which is not a limiting factor for the proposed multi-channel transformer approach. See supplementary material for qualitative translation examples from our best multi-articulatory translation model.

\section{Conclusion}
\label{sec:conc}
This paper presented a novel approach to Neural Machine Translation in the context of sign language. Our novel multi-channel transformer architecture allows both the inter and intra contextual relationship between different asynchronous channels to be modelled within the transformer network itself. Experiments on RWTH-PHOENIX-Weather-2014T dataset demonstrate the approach achieves \textit{on par} or competitive performance against the state-of-the-art. More importantly, we overcome the reliance on gloss information which underpins other state-of-the-art approaches. Now we have broken the dependency upon gloss information, future work will be to scale learning to larger dataset where gloss information is not available, such as broadcast footage.

\section*{Acknowledgements}
This work received funding from the SNSF Sinergia project `SMILE' (CRSII2\_160811), the European Union's Horizon2020 research and innovation programme under grant agreement no. 762021 `Content4All' and the EPSRC project `ExTOL' (EP/R03298X/1). This work reflects only the author’s view and the Commission is not responsible for any use that may be made of the information it contains. We would also like to thank NVIDIA Corporation for their GPU grant.

\clearpage
{\small
\bibliographystyle{splncs04}
\bibliography{Bib/action,Bib/camgoz,Bib/ctc,Bib/deeplearning,Bib/generative,Bib/gesture,Bib/misc,Bib/nmt,Bib/pose,Bib/seq2seq,Bib/sign,Bib/speech}
}
\end{document}


\pagestyle{headings}
\mainmatter
\def\ECCVSubNumber{12}  

\title{Multi-channel Transformers for\\Multi-articulatory Sign Language Translation: Supplementary Material} 

\titlerunning{Multi-channel Transformers for Multi-articulatory SLT}
%
\author{
Necati Cihan Camgoz\inst{1} \and
Oscar Koller\inst{2} \and
Simon Hadfield\inst{1} \and
Richard Bowden\inst{1}
}
\authorrunning{N. C. Camgoz et al.}
%
\institute{CVSSP, University of Surrey, UK, \{n.camgoz, s.hadfield, r.bowden\}@surrey.ac.uk,
\and
Microsoft, Munich, Germany, oscar.koller@microsoft.com
}

\maketitle

In this supplementary material, we report our quantitative experiment results for finding the best channel feature and word embedding setup. We also share qualitative translation samples produced by our best performing Multi-channel transformer model.

\section{Channel Feature and Word Embeddings}

In this preliminary set of experiments we investigate different ways to embed \ac{cnn} based channel features and one-hot word vectors. Machine translation orientated transformer implementations either use pretrained word embeddings or train a linear projection layer from scratch. Although not stated in the original paper \cite{vaswani2017attention}, the official transformer implementation also utilizes \textit{embedding scaling}, where the projected word representations are multiplied by a constant which is the square root of the hidden size\footnote{\url{github.com/tensorflow/models/blob/master/official/nlp/transformer/}}. 

Compared to the one-hot vectors which have a constant scale between $[0-1]$, \ac{cnn} features can have an arbitrary scale. To see how important the input scale is and to examine the effects of the different embedding setups, we initially trained translation networks that only used hand channel features as input. Each network has two layers, with a hidden size of 64 and 128 position-wise feed forward units. 

As can been in the first row of Table~\ref{multitrans:tbl:embed}, the translation performance degrades drastically when we apply the commonly used embedding scaling on either of the embeddings. We have experimentally found that transformer networks are extremely sensitive to input scale and this is substantiated by these results. Thus, in our next set of experiments we investigate ways to normalize inputs to control their scale. To do so, we utilize batch normalization~\cite{ioffe2015batch} and soft-sign activation~\cite{nwankpa2018activation}. While batch normalization scales the inputs between $[-3,3]$ it has also been shown to improve convergence rate. On the other hand, soft-sign activation scales the inputs between $[-1,1]$ while also enhancing the representation capability of the embedding due to its non-linear nature. 

Individually, both batch norm and soft-sign significantly improve the translation performance when applied to the projected \ac{cnn} features (see second and third rows of Table~\ref{multitrans:tbl:embed}). We then investigate their combined use on both \ac{cnn} features and word embeddings. Although there were several comparable setups, we concur that the joint application of batch norm and soft-sign only on the \ac{cnn} features yield the most stable and balanced performance in terms of development and test set for BLEU-4 and ROUGE scores. We believe this is due to the already scaled nature of the word embeddings and the additional stability and non-linearity introduced by applying soft-sign and batch norm to the projected \ac{cnn} features. Therefore, we utilize this embedding setup for our experiments in the main manuscript.

\begin{table}[h!]
\centering
\caption{Effects of using different embedding setups on hand channel features to spoken language translation performance.}
\label{multitrans:tbl:embed}
\resizebox{\linewidth}{!}{
\begin{tabular}{ccc|ccc|cc|cc|cc|cc}
\multicolumn{6}{c|}{} & \multicolumn{4}{c|}{Dev Set} & \multicolumn{4}{c}{Test Set} \\
\multicolumn{3}{c|}{Feature Embedding} & \multicolumn{3}{c|}{Word Embedding} & \multicolumn{2}{c|}{BLEU-4} &\multicolumn{2}{c|}{ROUGE} & \multicolumn{2}{c|}{BLEU-4} &\multicolumn{2}{c}{ROUGE} \\ \hline
Scaling&BatchNorm& SoftSign& Scaling&Batch Norm& Soft-Sign& Best  & mean $\pm$ std   & -     &  mean $\pm$ std   & -    &  mean $\pm$ std   & -    &  mean $\pm$ std  \\ \hline \hline 
\crx & -    & -    & \crx & -    & -    & \textbf{15.46} & \textbf{15.08} $\pm$ 0.25 & \textbf{40.57} & \textbf{39.39} $\pm$ 0.55 & \textbf{15.98} & \textbf{15.23} $\pm$ 0.54 & \textbf{39.97} & \textbf{39.44} $\pm$ 0.80 \\ 
\cbt & -    & -    & \crx & -    & -    & 13.96 & 13.39 $\pm$ 0.36 & 37.51 & 37.07 $\pm$ 0.33 & 14.29 & 13.77 $\pm$ 0.47 & 37.91 & 37.54 $\pm$ 0.60 \\ 
\crx & -    & -    & \cbt & -    & -    & 14.54 & 14.20 $\pm$ 0.26 & 37.86 & 38.02 $\pm$ 0.70 & 14.80 & 14.58 $\pm$ 0.46 & 38.35 & 38.38 $\pm$ 0.72 \\ 
\cbt & -    & -    & \cbt & -    & -    & 13.52 & 12.88 $\pm$ 0.36 & 36.93 & 36.14 $\pm$ 0.66 & 13.99 & 13.29 $\pm$ 0.53 & 37.72 & 36.58 $\pm$ 0.81 \\ 
\hline
\hline
\ckx & \cbt & -    & \ckx & \crx & -    &\textbf{16.35} &\textbf{15.64} $\pm$ 0.46 &\textbf{41.30} &\textbf{40.41} $\pm$ 0.55 &\textbf{16.09} &\textbf{15.60} $\pm$ 0.50 &\textbf{40.89} &\textbf{40.15} $\pm$ 0.70 \\ 
\ckx & \crx & -    & \ckx & \cbt & -    & 14.49 & 14.07 $\pm$ 0.25 & 37.96 & 37.86 $\pm$ 0.58 & 14.72 & 14.33 $\pm$ 0.38 & 37.49 & 37.53 $\pm$ 0.43 \\ 
\ckx & \cbt & -    & \ckx & \cbt & -    & 15.17 & 14.69 $\pm$ 0.30 & 39.92 & 39.10 $\pm$ 0.56 & 15.50 & 14.99 $\pm$ 0.64 & 39.83 & 39.26 $\pm$ 0.81 \\ 
\hline
\ckx & -    & \cbt & \ckx & -    & \crx &\textbf{16.40} &\textbf{15.74} $\pm$ 0.55 &\textbf{41.90} &\textbf{40.73} $\pm$ 0.63 & 14.95 &\textbf{15.63} $\pm$ 0.48 & 39.31 & 39.93 $\pm$ 0.57 \\ 
\ckx & -    & \crx & \ckx & -    & \cbt & 15.75 & 15.10 $\pm$ 0.35 & 39.72 & 39.47 $\pm$ 0.46 &\textbf{16.33} & 15.41 $\pm$ 0.55 & 40.74 & 39.76 $\pm$ 0.67 \\ 
\ckx & -    & \cbt & \ckx & -    & \cbt & 15.98 & 15.50 $\pm$ 0.35 & 41.02 & 40.24 $\pm$ 0.51 & 15.64 & 15.49 $\pm$ 0.48 &\textbf{40.79} &\textbf{40.02} $\pm$ 0.59 \\ 
\hline
\hline
\ckx & \cbt & \cbt & \ckx & \crx & \crx &\textbf{16.44} &\textbf{15.70} $\pm$ 0.41 & 40.79 & 40.45 $\pm$ 0.64 &\textbf{16.18} & 15.63 $\pm$ 0.65 &\textbf{40.62} & 40.07 $\pm$ 0.80 \\ 
\ckx & \crx & \crx & \ckx & \cbt & \cbt & 15.15 & 14.18 $\pm$ 0.42 & 39.30 & 37.78 $\pm$ 0.77 & 15.14 & 14.30 $\pm$ 0.53 & 38.41 & 37.53 $\pm$ 0.82 \\ 
\ckx & \cbt & \cbt & \ckx & \cbt & \cbt & 15.59 & 15.04 $\pm$ 0.33 & 39.83 & 39.40 $\pm$ 0.46 & 14.85 & 14.99 $\pm$ 0.51 & 39.26 & 39.30 $\pm$ 0.65 \\ 
\ckx & \cbt & \cbt & \ckx & \crx & \cbt & 15.98 & 15.62 $\pm$ 0.21 &\textbf{41.63} &\textbf{40.63} $\pm$ 0.62 & 15.97 &\textbf{15.65} $\pm$ 0.49 & 40.54 &\textbf{40.24} $\pm$ 0.74 \\ 
\ckx & \crx & \cbt & \ckx & \cbt & \cbt & 15.09 & 14.70 $\pm$ 0.26 & 39.36 & 38.95 $\pm$ 0.25 & 14.70 & 14.80 $\pm$ 0.49 & 39.38 & 38.87 $\pm$ 0.67 \\ 
\hline
\end{tabular}
 }
\end{table}

\section{Qualitative Examples}
\label{multitrans:sec:qual}

In this section we share translation examples generated by our best performing model. As the ground truth spoken language annotations of the \ac{ph14t} dataset are in German, we share both the original German translations and their equivalent word-by-word translations in English. As can be seen in Table ~\ref{multitrans:tbl:qual:results}, we also categorize the results into three categories, namely \emph{Good}, \emph{Mediocre} and \emph{Poor} translations, to give further insight to the reader on the limitations of the current approach.

We categorize translations as \emph{Good} or \emph{Mediocre} when the produced sentences convey the same or similar information as the reference sentences. These examples follow the standard grammar with few exceptions. We classify translations as \emph{Poor} when the model fails to understand and translate the conveyed information in sign videos. Most of these examples contain repetitions. In some cases, the model is not able distinguish some sign glosses from another, such as named entities like locations or numbers which occur in limited contexts in the training data. One way to address this issue might be to utilize pretrained spoken language models to improve the produced translations. 

\begin{table}[ph!]
\caption{Spoken language translations produced by our best {Multi-Channel Transformer} model.}
\label{multitrans:tbl:qual:results}
\centering
    \resizebox{\columnwidth}{!}{
    \begin{tabular}{|rl|}
    \multicolumn{2}{l}{Good Translations:}\\
    \hline
    Reference:  & und nun die wettervorhersage für morgen dienstag den ersten februar . \\
    ~           & ( and now the weather forecast for tomorrow tuesday the first of february . ) \\
    Ours:       & und nun die wettervorhersage für morgen dienstag den ersten februar . \\
    ~           & ( and now the weather forecast for tomorrow tuesday the first of february . ) \\
    \hline
    Reference:  & der sorgt wieder für wolken die regen im bergland auch schnee bringen . \\
    ~           & ( it provides clouds again and the rain in the mountains also brings snow . ) \\
    Ours:       & im übrigen land fällt gebietsweise regen im bergland auch schnee . \\
    ~           & ( in the rest of the country there is rain in the mountains and snow in some areas. ) \\
    \hline
    Reference:  & die neue woche beginnt wechselhaft und kühler . \\
    ~           & ( the next week starts variable and colder . ) \\
    Ours:       & auch am montag wechselhaft und deutlich kühler . \\
    ~           & ( also on monday variable and significantly colder . ) \\
    \hline
    \multicolumn{2}{l}{Mediocre Translations:}\\
    \hline
    Reference:  & ab sonntag wird es wieder milder dabei gibt es viele wolken zeitweise fällt regen im nordwesten windig . \\
    ~           & (from sunday on it will be more mild with many clouds partly rain in the northwest windy .) \\
    Ours:       & am sonntag mehr wolken als sonne hier und da regen im westen ist es windig . \\
    ~           & (on sunday more clouds than sun from time to time rain in the west windy .) \\
    \hline
    Reference:  & im osten und südosten auch schnee oder schneeregen . \\
    ~           & ( in the east and south-east also snow or sleet . ) \\
    Ours:       & im südosten schnee oder schnee . \\
    ~           & ( snow or snow in the south-east . ) \\
    \hline
    Reference:  & westlich des rheins und im nordosten bleibt es meist trocken . \\
    ~           & ( west of the rhine and in the northeast it remains mostly dry . ) \\
    Ours:       & im westen und südwesten bleibt es noch im nordosten trocken . \\
    ~           & ( in the west and southwest it remains still dry in the northeast . ) \\
    \hline
    \multicolumn{2}{l}{Poor Translations:}\\
    \hline
    Reference:  & deutschland liegt morgen unter hochdruckeinfluss der die wolken weitgehend vertreibt . \\
    ~           & ( germany will be under the influence of high pressure tomorrow which will largely dispel the clouds . ) \\
    Ours:       & deutschland liegt morgen über deutschland nach deutschland . \\
    ~           & ( germany is tomorrow over germany to germany . ) \\
    \hline
    Reference:  & am freitag insgesamt viele wolken die regen bringen . \\
    ~           & ( on friday overall many clouds bringing rain . ) \\
    Ours:       & am freitag gibt es am freitag viele wolken . \\
    ~           & ( on friday there are on friday many clouds ) \\
    \hline
    Reference:  & dazu weht ein starker wind vor allen dingen wieder über vorpommern aus südost . \\
    ~           & ( in addition a strong wind blows before all things again over vorpommern from southeast . ) \\
    Ours:       & es weht ein kräftiger nordostwind . \\
    ~           & ( a strong north-easterly wind is blowing . ) \\
    \hline

    \end{tabular}}
\end{table}

{\small
\bibliographystyle{splncs04}
\bibliography{Bib/action,Bib/camgoz,Bib/ctc,Bib/deeplearning,Bib/generative,Bib/gesture,Bib/misc,Bib/nmt,Bib/pose,Bib/seq2seq,Bib/sign,Bib/speech}
}